# Multiple-Channel Real Time Filtering for a Myoelectric Prosthetic Hand-Arm Robot System

Weibang Bai[1], Yinlai Jiang[2], Hiroshi Yokoi[2]

[1] Shangai Jiao Tong University, Shanghai, China.

[2] The University of Electro-Communications, Tokyo, Japan.

**Abstract**

On the base of the developed master-slave prosthetic hand-arm robot system, which is controlled mainly based on signals obtained from bending sensors fixed on the data glove, the first idea deduced was to develop and add a multi-dimensional filter into the original control system to make the control signals cleaner and more stable at real time. By going further, a second new idea was also proposed to predict new control information based on the combination of a new algorithm and prediction control theory. In order to fulfill the first idea properly, the possible methods to process data in real time, the different ways to produce Gaussian distributed random data, the way to combine the new algorithm with the previous complex program project, and the way to simplify and reduce the running time of the algorithm to maintain the high efficiency, the real time processing with multiple channels of the sensory system and the real-time performance of the control system were researched. Eventually, the experiment on the same provided robot system gives the results of the first idea and shows the improved performance of the filter comparing with the original control method.

## 1. Background

We have developed the master-slave hand-arm robotic system in the Lab lead by Prof. Hiroshi Yokoi in The University of Electro-Communications (UEC), Tokyo, Japan. The overall setup of the robot system is shown in Fig.1, it is a multiple-DoFs myoelectric prosthetic hand system developed with an interactive-tendon driven mechanism and controlled with master-slave mapping [1]. The detailed power transmission and actuation architecture using the tendon-driven mechanism is shown in Fig.2.

On the master side, when the human operator moves his hand and arm, the joint changing information of the arm and each finger would be collected from the bending sensors inside the data glove and other sensors installed near each joint. Currently, the original data obtained from the sensors are sent to the driving motors after two processing steps: the first step is linearization, and the second one is to transfer bending angles into motor commands. The principle of the control signals in the master-slave robot system can be seen in the Fig.3.

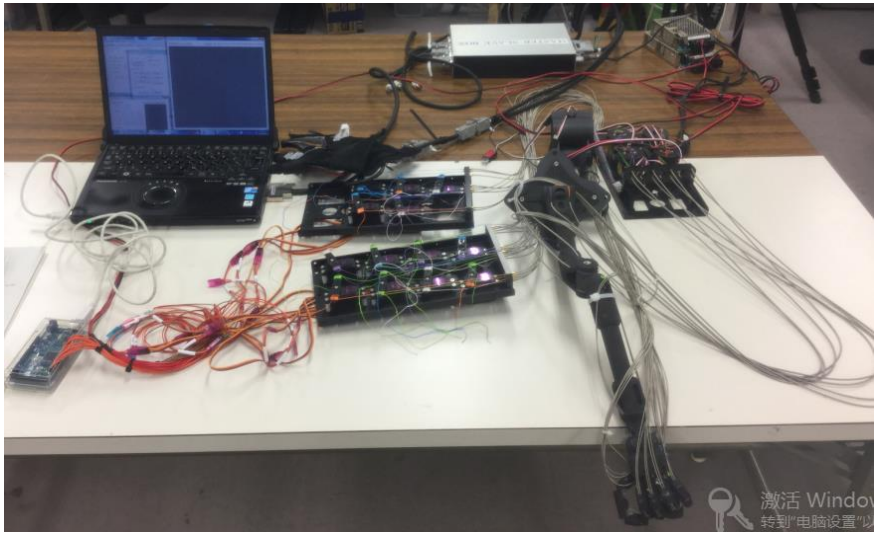

Fig.1 The master-slave prosthetic hand-arm robot system in the Lab

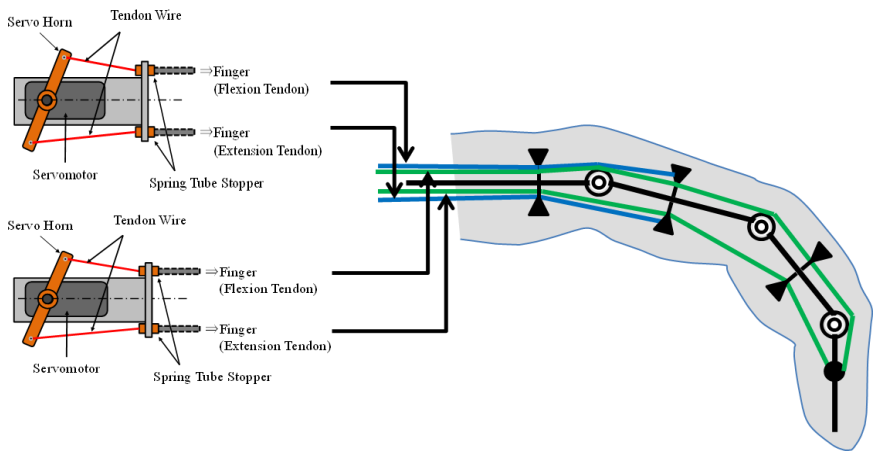

Fig.2 The actuation architecture of the finger in the prosthetic hand

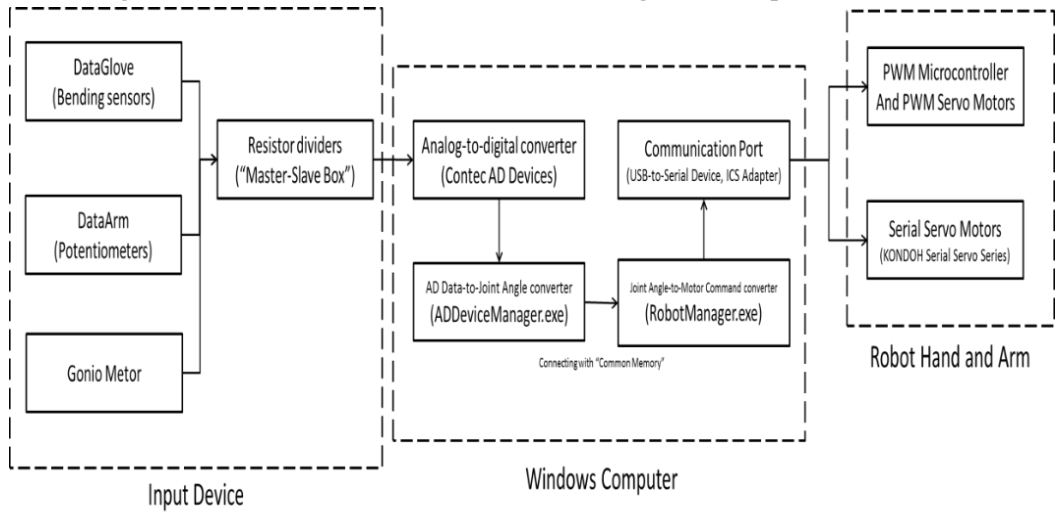

Fig.3 The principle of the control signals of the master-slave robot system

After having a deep look into the control method, I found that the data obtained from the data glove made by the lab is using many bending sensors and it is directly linearized to calculate the control commands of the driving motors. What's more, when outputting and visualizing the original data from the glove, I found that the data was very noisy, as shown in Fig.4.

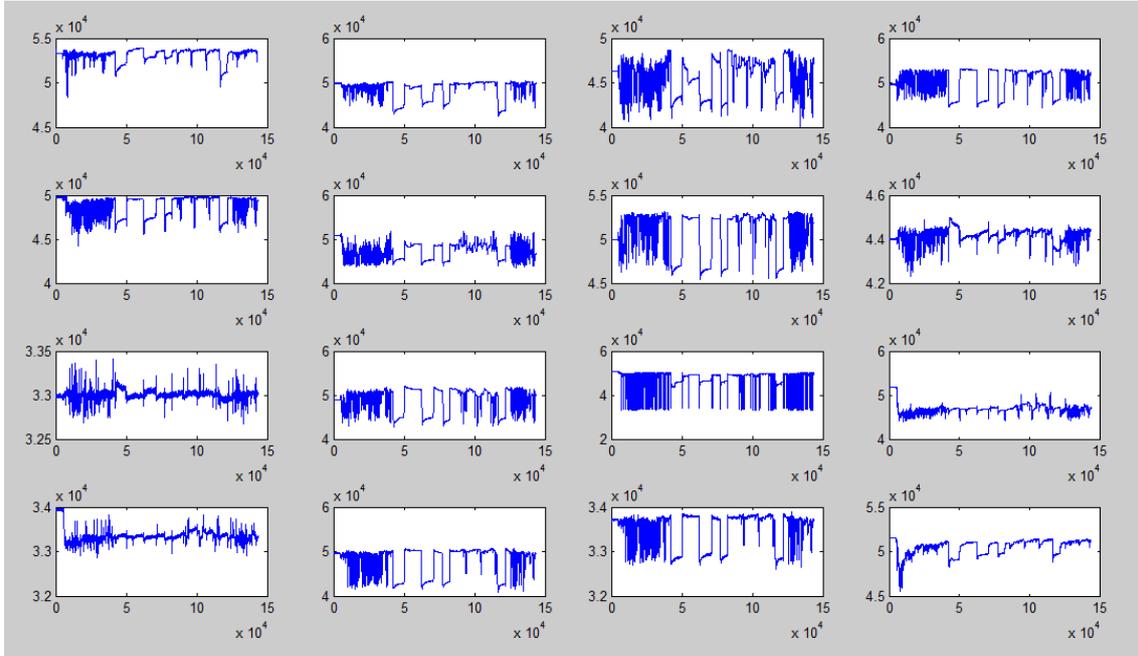

Fig.4 Original noisy data obtained from the glove

## 2. Concepts and development

### 2.1 Method to process data in real time

If there is a filter or an algorithm that can reduce or compensate the noise of the ongoing generating data in real time, we can get better control performance of the master-slave system. Or even the simple and basic mean filter can be tried. But the results of the test with this simple filter were not as good and it reduced the real-time performance seriously. So it needs a new different filter or algorithm that can process the obtained data simultaneously. In the beginning, I tried to develop a new algorithm by myself and also I searched many papers. While after having read many materials, I found that the kalman filter can be working for this first idea, but the thing is to adapt it and make it appliable for the multiple channels in this sensory system of the robot.

### 2.2 Kalman Filter algorithm

After reviewing the literature, the famous Kalman filter was found as it can process the data that meets the basic idea and has been widely used in many different fields. Here is the basic concept of the Kalman filter algorithm in Fig.5, which is referred to [2].

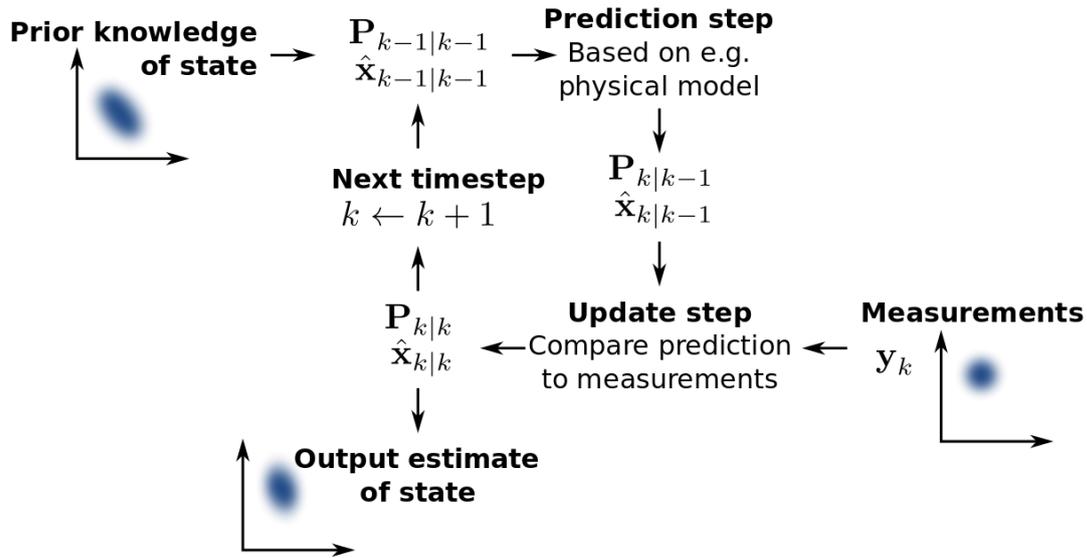

Fig.5 The basic concept of Kalman filtering [2]

The Kalman filter keeps track of the estimated state of the system and the variance or uncertainty of the estimate. The estimate is updated using a state transition model and measurements.

## 2.3 Development of the Kalman algorithm
### 2.3.1 Analyze the measurement system in the robot system

A Linear Dynamical System is a partially observed stochastic process with linear dynamics and linear observations, both subject to Gaussian noise. Because the bending sensors are detecting the angles of each joint of operator's fingers, we need care about the model of the measurement system. For a proper application of the filter, we need to set up a linear model or linearize the original model of the measure system.

A more easy and usual way is to simply define the model according to the classic method. And we can get different equations for different situations. As can be described like this:

For a single joint:

$$\begin{bmatrix} q(k+1) \\ w(k+1) \end{bmatrix} = \begin{bmatrix} 1 & 1 \\ 0 & 1 \end{bmatrix} \begin{bmatrix} q(k) \\ w(k) \end{bmatrix} + \begin{bmatrix} 0.5 \\ 1 \end{bmatrix} * alfa + \begin{bmatrix} v_1(k) \\ v_2(k) \end{bmatrix}$$

$$z(k) = \begin{bmatrix} 1 & 0 \end{bmatrix} \begin{bmatrix} q(k) \\ w(k) \end{bmatrix} + n(k)$$

For multiple joints:

$$\begin{bmatrix} Q(k+1)_{n\times 1} \\ W(k+1)_{n\times 1} \end{bmatrix} = \begin{bmatrix} A11_{n\times n} & A12_{n\times n} \\ A21_{n\times n} & A22_{n\times n} \end{bmatrix} \begin{bmatrix} Q(k)_{n\times 1} \\ W(k)_{n\times 1} \end{bmatrix} + \begin{bmatrix} D_{n\times n} \\ E_{n\times n} \end{bmatrix} * Ac_{n\times 1} + W_N(k)_{n\times 1}$$

$$Z(k)_{n\times n} = \begin{bmatrix} H_{n\times n} & 0 \end{bmatrix} \begin{bmatrix} Q(k)_{n\times 1} \\ W(k)_{n\times 1} \end{bmatrix} + V_N(k)_{n\times 1}$$

where $v_1(k)$, $v_2(k)$ are Gaussian random noise, $W_N(k)_{n \times 1}$、 $V_N(k)_{n \times 1}$ are matrix with elements of Gaussian random noise.

### 2.3.2 The definition of the key parameters in the kalman algorithm

There are few key parameters in the algorithm and we need to define and calculate each of them. The equations of each of them are list as follows:

$$\hat{x}_j^- = A\hat{x}_{j-1} + Bu_j$$

$$\hat{x}_j = \hat{x}_j^- + K_j\left(z_j - H\hat{x}_j^-\right)$$

$$P_j^- = E\{e_j^- e_j^{-T}\} = E\left\{\left(x_j - \hat{x}_j^-\right)\left(x_j - \hat{x}_j^-\right)^T\right\}$$

$$P_j = E\{e_j^- e_j^{-T}\} = E\left\{\left(x_j - \hat{x}_j\right)\left(x_j - \hat{x}_j\right)^T\right\}$$

$$\frac{\partial P_j}{\partial K_j} = \frac{\partial E\left\{\left(x_j - \hat{x}_j\right)\left(x_j - \hat{x}_j\right)^T\right\}}{\partial K_j} = 0$$

$$K_j = \frac{P_j H^T}{H P_j H^T + R}$$

$$P_j^- = A P_{j-1} A^T + Q$$

$$P_j = (I - K_j H) P_j^-$$

By following the equations above we can get the key parameters of the algorithm and we can update them step by step.

### 2.3.3 Generate the Gaussian distributed data

The linearized model has taken the random noise into consideration, so we need to provide the noise in our realization algorithm. And before that works, the methods of generating the needed need to be researched.

### 2.3.3.1 Box-Muller method

The Box–Muller transform [3] (by George Edward Pelham Box and Mervin Edgar Muller 1958) is a pseudo-random number sampling method for generating pairs of independent, standard, normally distributed (zero expectation, unit variance) random numbers, given a source of uniformly distributed random numbers. Suppose U1 and U2 are independent random variables that are uniformly distributed in the interval (0, 1]. In the basic form (the other one is polar form), let

$$Z_0 = R\cos(\Theta) = \sqrt{-2\ln U_1}\cos(2\pi U_2)$$

$$Z_1 = R\sin(\Theta) = \sqrt{-2\ln U_1}\sin(2\pi U_2).$$

Then Z0 and Z1 are independent random variables with a standard normal distribution. The Fig.6 [3] taken from Wikipedia explains the free encyclopedia: Visualization of the Box-Muller transform — the colored points in the unit square (u1, u2), drawn as circles, are mapped to a 2D Gaussian (z0, z1), drawn as crosses. The plots at the margins are the probability distribution functions of z0 and z1. Note that z0 and z1 are unbounded; they appear to be in [-3,3] due to the choice of the illustrated points.

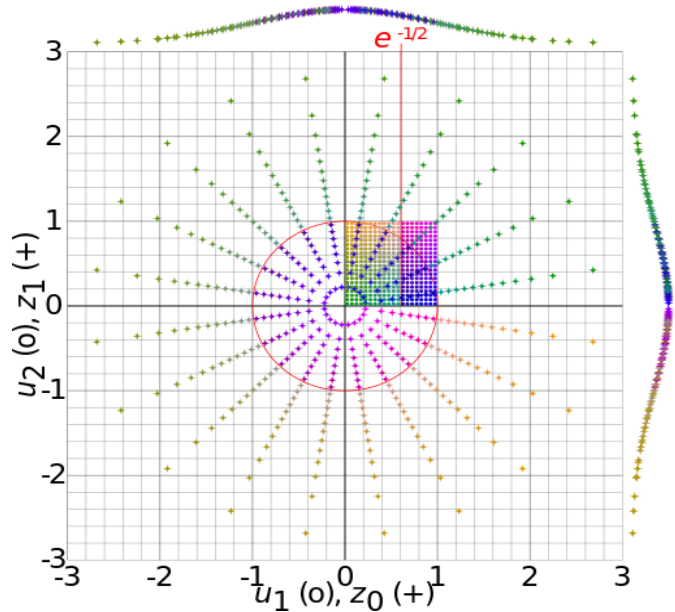

Fig.6 Visualization of the Box-Muller transform (from Wikipedia)[3]

After tried the algorithm based on this method in Visual Studio 2010, the result of the data can be drawn in Matlab, as shown in Fig.7.

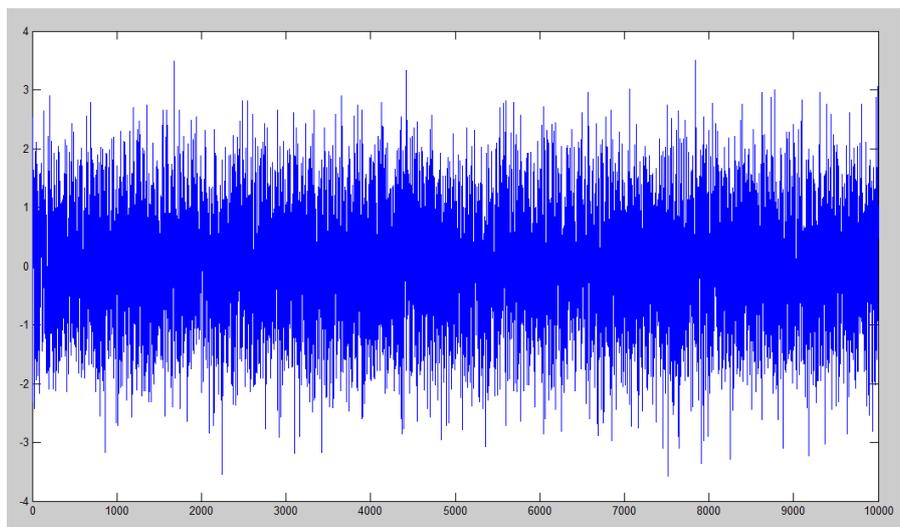

Fig.7 The output of the random data generated using the Box-Muller method

### 2.3.3.2 Galton board method

The Galton box, is a device invented by Sir Francis Galton to demonstrate the central limit theorem [4,5], in particular that the normal distribution is approximate to the binomial distribution. There is a vertical board with interleaved rows of pins, and balls are dropped from the top, and bounce left and right as they hit the pins. The height of ball columns in the bins approximates a bell curve. The basic idea of the Galton board method is shown in Fig.8, which is referred from the website [4].

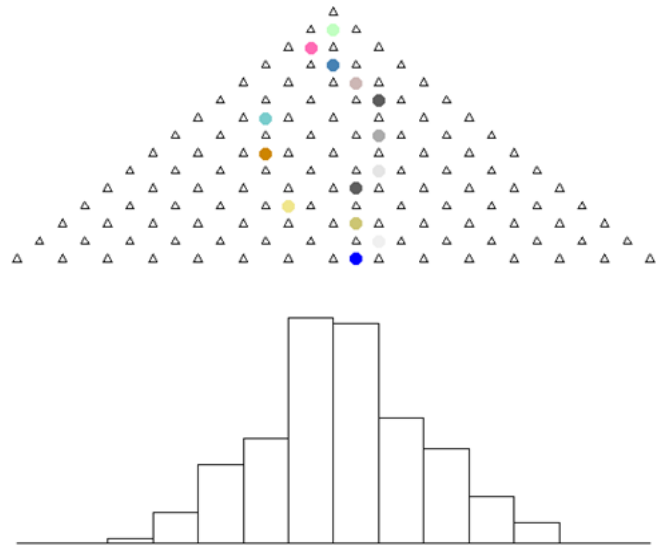

Fig.8 the basic idea of the Galton board method

And this method was also tested in Visual Studio 2010, and the outcome of the generated data was draw in Excel, as shown in Fig.9. The results proved that the random numbers generated using this method are following Gaussian distributed.

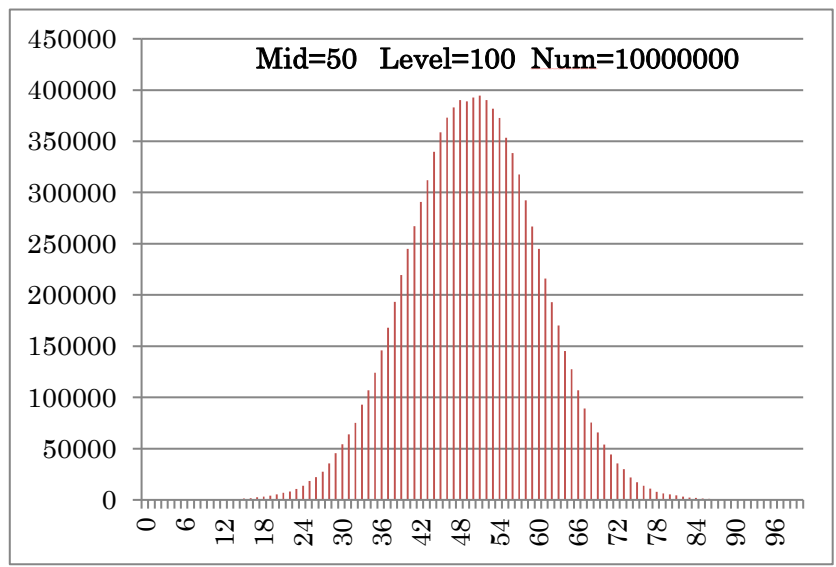

Fig.9 the distribution of the random data generated using the Galton board method

### 2.3.4 Implement of the Kalman filter

At first, the algorithm was simply developed in Matlab, which is very convenient to develop different algorithms with a lot of mathematic tools. After tested in Matlab, it is developed in C++ using Visual Studio. Because the algorithm involves a lot calculation about matrix of which the size is changing according to the channels chosen in the sensory system, it is hard to exactly realize each step of the calculation among matrixes. But there are many different templates and libraries developed for C++ developers by different researchers around the world. Thus it is very convenient to develop different kinds of professional algorithms in C++. And Eigen is chosen to help developing them successfully.

After the realization of the algorithm in C++, real tests can be completed. The first step of the test is to get the original data from the sensors. In order to have a look at the effect of the algorithm applied in the real robot system, the first 6000 time steps of the data output from one of the bending sensors was firstly processed. The result is shown in Fig.10.

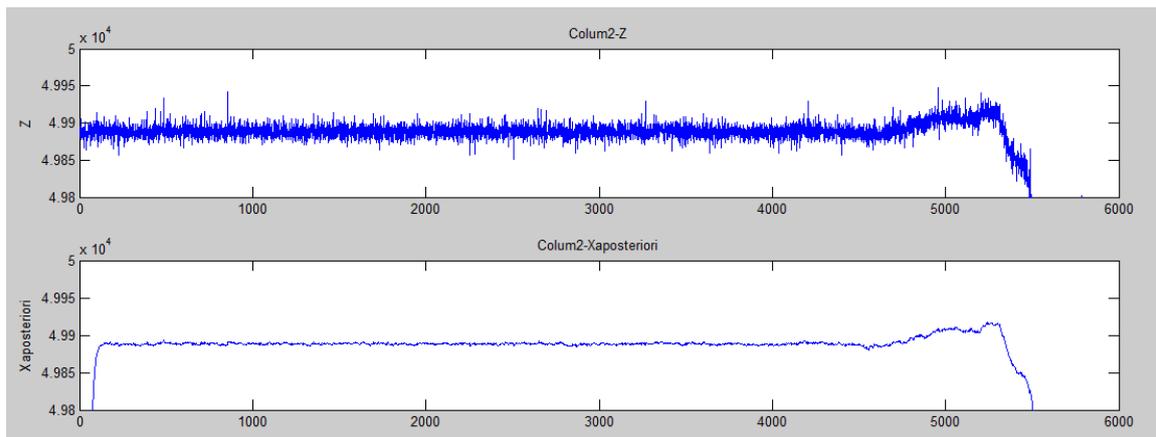

Fig.10 the first result of the algorithm applied in the real data test

We can see that the result (the curve Xaposteriori) was much better than the original data (the curve Z). But this test was not carried out in real time with the robot system as proposed at first. So the next step is to combine the filter into the original control system.

### 2.4 Configuration of the variance-covariance matrixes for multiple channels

In the prosthetic hand-arm robot system, the sensory system is in high dimensions, and even the dimension can be dynamic when operating as the user can choose them. And all these channels need to be processed in real time, So multiple-channel real Time filtering is of great importance for this myoelectric prosthetic hand-arm robot system. The model has taken the random noise into consideration, and the variance-covariance matrixes of Q and R should be reconfigured and optimized to make the filter works better. For the measurement part of the model, after having analyzed the sampling data obtained from 18 channels, the variance-covariance matrix of R in 18 dimensions was calculated. R is a diagonal matrix and the result is:

R = diag [21.18   12.52   10.62   9.86    26.94   10.08
         9.71    17.58   6.05    9.91    47.19   11.04
         7.56    23.94   26.82   28.82   27.69   31.49];

i.e. $R \in \mathcal{R}^{18 \times 18}$. And for the system part of the model, the matrix Q I just simply make it like: Q = I, where I is unit matrix of the same dimension with the system model.

## 3. Method Integration

After the implementation of the filter, we need to combine the filter into the original control system. Before combining the new algorithm into the control system, the original control system should be investigated, and also several principles for the combination can be figured out: Firstly, not change the structure the control system; Secondly, keep the real time performance and assure the data sampling rate should be higher than 1000Hz, making the increase of the event time as less as possible.

The real-time performance is very important in a master-slave system especially for a master-slave surgical prosthetic robotic system. So the data sampling rate and the event time are of great importance. To release that, a lot of time was spent to simplify the new algorithm and to connect to the previous system more properly. Finally it was achieved after more than 4 different versions of the program with many problems appeared, and the sampling rate is at 1000Hz and the event time is less than 45ms.

## 4. Experiment, Results, and Conclusion

Having combined successfully after a long time and a lot of tries, the experiments can be completed to test the real-time master side control data processing with filtering. The conditional parameters of the first real-time test are listed as follows:

- Sampling rate: 1000Hz;
- Event rate: 0.02s;
- Sensor channels: 18;
- Model dimensions: 18;
- Input dimensions: 18;
- Output dimensions: 18;
- Data amount: 94096 x 18;
- Master motion type: random;
- Data processing: real time;

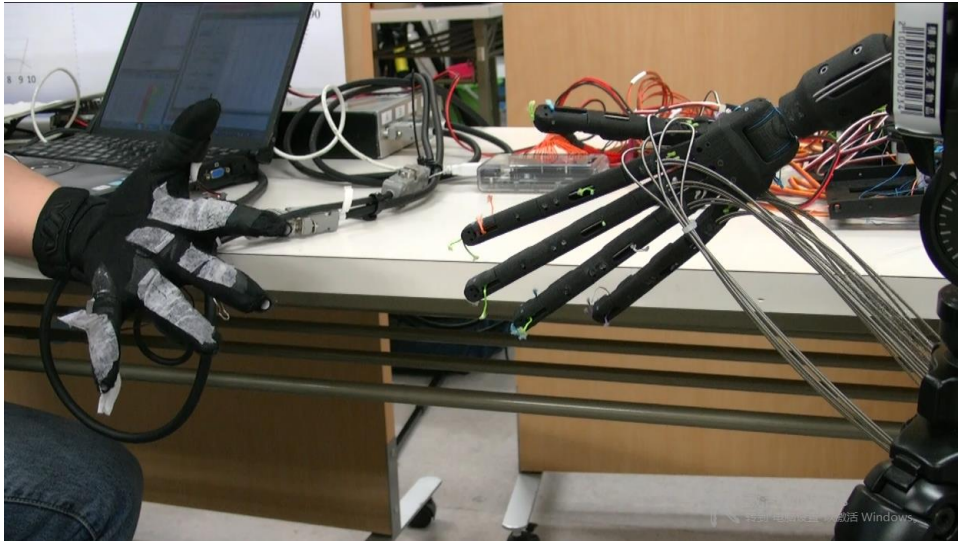

Fig.11 Test on the prosthetic hand arm robot system

Under the condition mentioned above, here got the results of the data that was processed simultaneously during the first real test, as shown in Fig.11. Considering the total channel number was 18, and there is no need to list all of them for the reason of the space, I just casually list few channels as representative. The figures of channel 1, channel 5, channel 8, channel 10, channel 13, and channel 16 are list as follows in Fig.12.

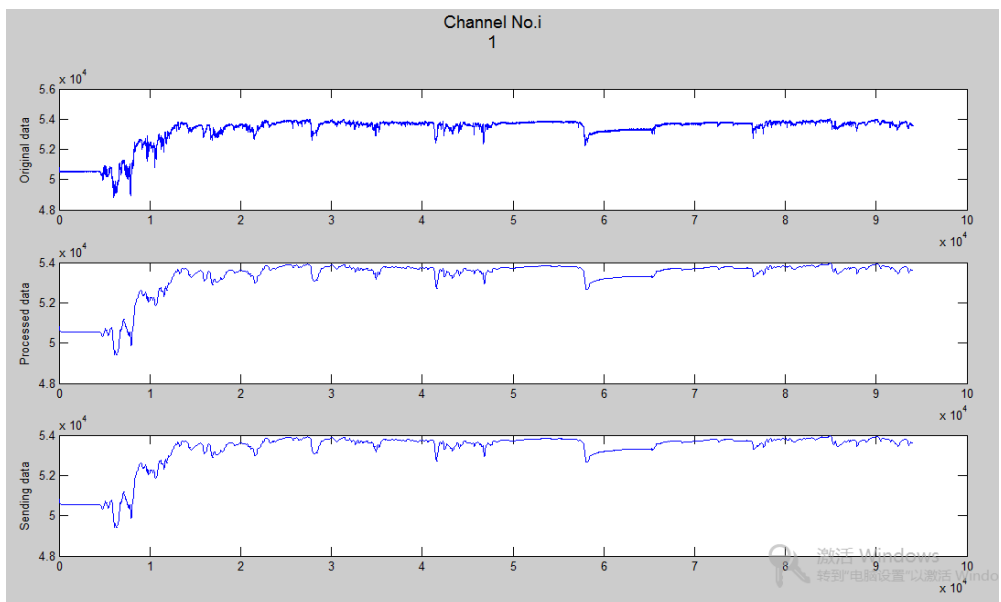

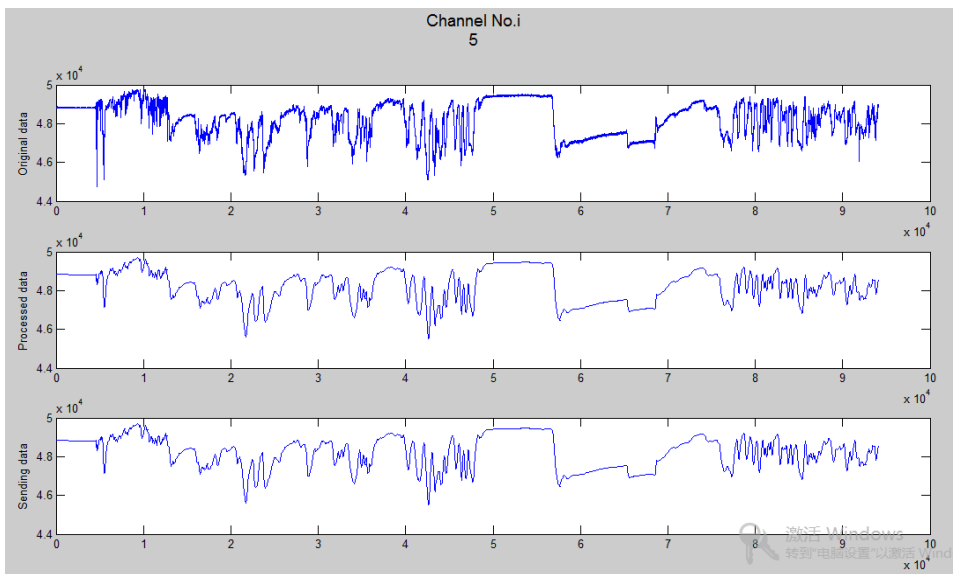

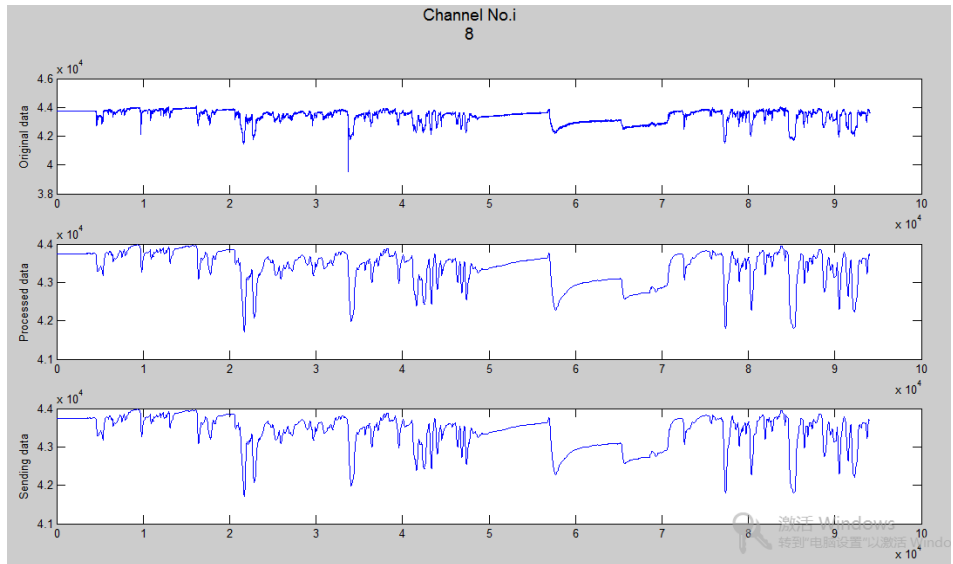

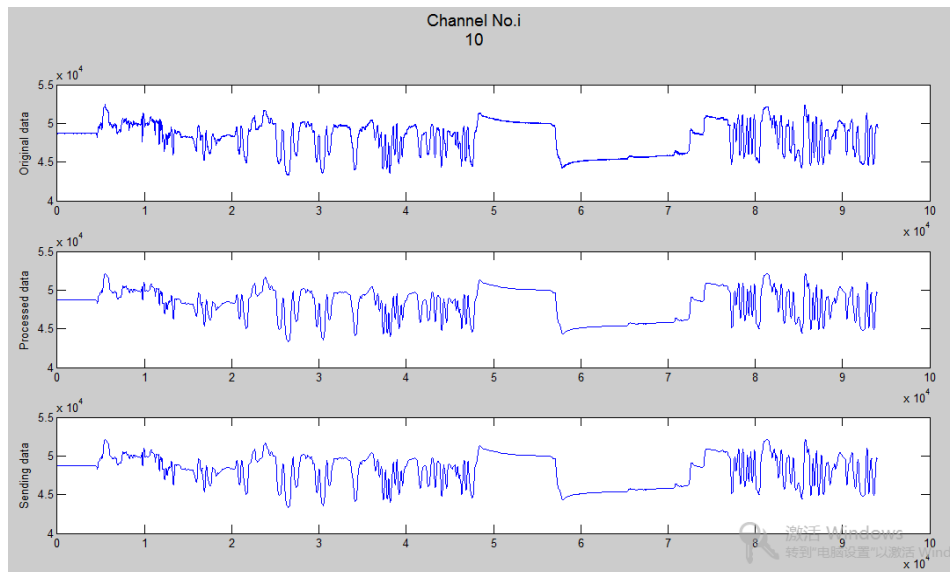

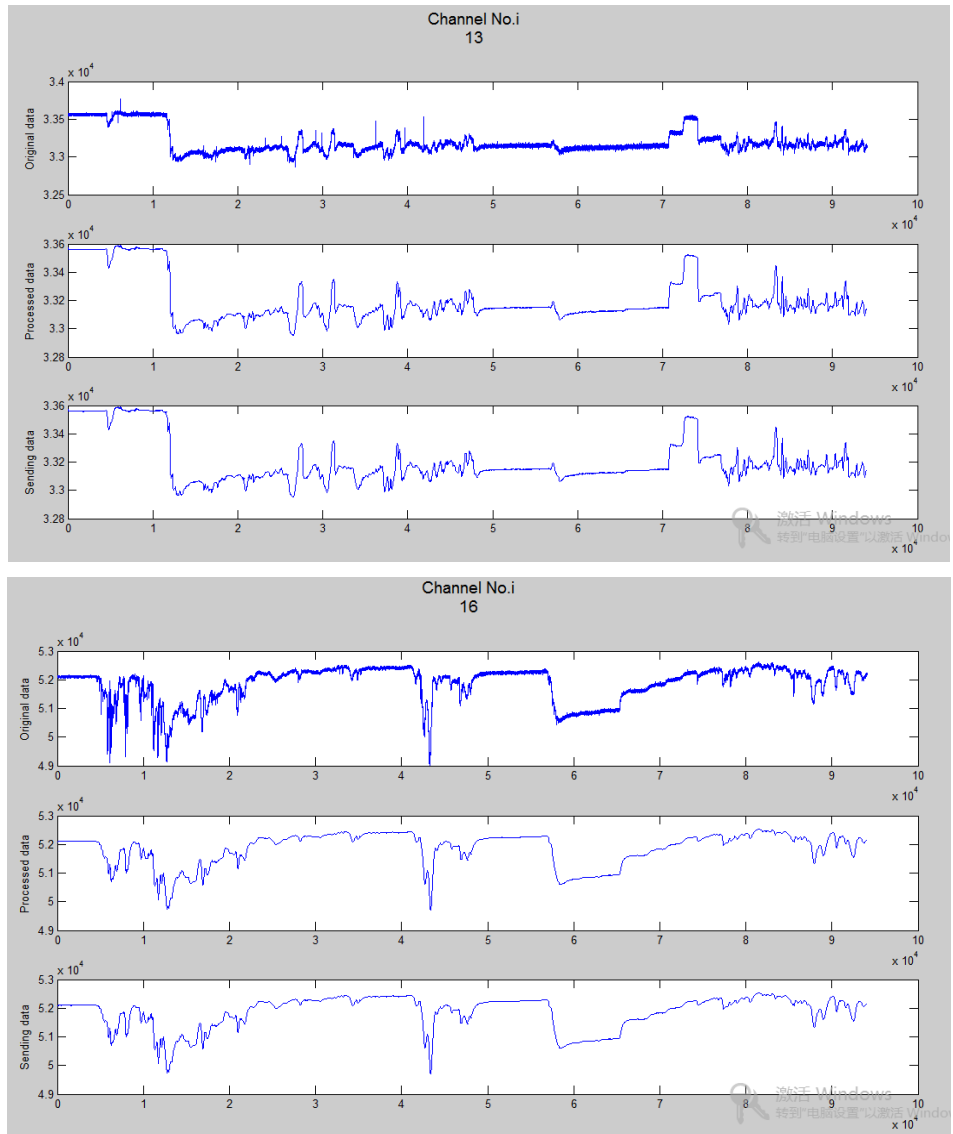

Fig.12 The listed results in different channels of the test

Each figure above in Fig.12 contains 3 curves for each channel, the first one represents the original data, the second one represents the processed data, and the last one represents the data which was sending to the driving motor during the real test. And because the data types defined are different between the original system and the new algorithm, the sending data was transferred from float to integer as the original control system used to control the driving motors.

As the results shows, most of the channels got a much better results after processed by the algorithm comparing with the original data. If we make a study of the results in detail, we can find that there are much less local jumps in the processed data curve than original ones. Consequently, each finger of the robot hand moved in a more quiet and more clean way. And the effect is more apparent especially when the motion is more slowly. At the same time, we can find the noise of the driving motors especially the frequency of the noise is much more less than the original control system. And the more precise the servo motor is, the better the effect would be.

Because the control system is not a closed loop system, the exact data directly is hard to obtain. And it is also not easy to measure the difference between the two control results because no proper measurement instruments can be used currently. But we got the video of experiments which can indicate the nuances.

## 5. Future work plan

At first, two ideas to get a different control result based on the original system were proposed. The first one was to process the data and to make the control signals less noisy in real time to get a better control performance of the master-slave hand-arm robot system. And this part has been achieved recently. But the progress was not that quick as planned because of some adjusting problems which need to be solved to make the new algorithm work properly with the original control system, and some other schedules also took a lot time. The second idea is still in the composing stage, and not enough time is left to realize and test it accordingly. So the next step should be to realize and test the second idea proposed in the beginning. The basic idea is to develop a new algorithm combining the prediction control theory to improve the response time and the master-slave following performance of the hand-arm system when the master side moves more quickly. Thus, this work can be complete in the future, and hope there will be a chance to test on the system again.

## Acknowledgement

This work is a brief report of the research work during the short-term research exchange program between University of Electro-Communications (UEC, Tokyo, Japan) and Shanghai Jiao Tong University (SJTU, Shanghai, China), which is sponsored by full scholarship from JASSO (Japan). Many thanks to the lovely lab members in Porf. Yorkoi Lab in UEC, especially to Prof. Hiroshi Yokoi, Dr. Yinlai Jiang, Ms. Yoshiko Yabuki, Mr. Tatsuya Sek, Ms. Xiaobei Jing, and Mr. Xu Yong. Submission date: 2015-01-28.